\documentclass{article}

\usepackage{arxiv}

\usepackage[utf8]{inputenc} 
\usepackage[T1]{fontenc}    
\usepackage{hyperref}       
\usepackage{url}            
\usepackage{booktabs}       
\usepackage{amsfonts}       
\usepackage{nicefrac}       
\usepackage{microtype}      
\usepackage{lipsum}
\usepackage{graphicx}
\usepackage{subfigure}
\usepackage{booktabs}
\usepackage{amsmath}
\usepackage{amssymb}
\usepackage{mathtools}
\usepackage{amsthm}
\usepackage{diagbox}
\usepackage{authblk}
\usepackage[capitalize,noabbrev]{cleveref}
\usepackage{multirow}
\usepackage{wrapfig}
\usepackage{bm}

\title{MCMARL: Parameterizing Value Function via Mixture of Categorical Distributions for Multi-Agent Reinforcement Learning}

\author[1]{Jian Zhao}
\author[1]{Mingyu Yang}
\author[1]{Youpeng Zhao}
\author[1]{Xunhan Hu}
\author[1]{Wengang Zhou}
\author[2]{Jiangcheng Zhu}
\author[1]{Houqiang Li}
\affil[1]{University of Science and Technology of China}
\affil[2]{Huawei Cloud}

\begin{document}
\maketitle

\begin{abstract}
In cooperative multi-agent tasks, a team of agents jointly interact with an environment by taking actions, receiving a team reward and observing the next state.
During the interactions, the uncertainty of environment and reward will inevitably induce stochasticity in the long-term returns and the randomness can be exacerbated with the increasing number of agents.
However, such randomness is ignored by most of the existing value-based multi-agent reinforcement learning (MARL) methods, which only model the expectation of Q-value for both individual agents and the team.
Compared to using the expectations of the long-term returns, it is preferable to directly model the stochasticity by estimating the returns through distributions.
With this motivation, this work proposes a novel value-based MARL framework from a distributional perspective, \emph{i.e.}, parameterizing value function via \underline{M}ixture of \underline{C}ategorical distributions for MARL. 
Specifically, we model both individual Q-values and global Q-value with categorical distribution.
To integrate categorical distributions, we define five basic operations on the distribution, which allow the generalization of expected value function factorization methods (\emph{e.g.}, VDN and QMIX) to their MCMARL variants.
We further prove that our MCMARL framework satisfies \emph{Distributional-Individual-Global-Max} (DIGM) principle with respect to the expectation of distribution, which guarantees the consistency between joint and individual greedy action selections in the global Q-value and individual Q-values.
Empirically, we evaluate MCMARL on both a stochastic matrix game and a challenging set of StarCraft II micromanagement tasks, showing the efficacy of our framework.
\end{abstract}

\section{Introduction}
Reinforcement learning (RL) aims to learn a mapping from the observation state to the action of an agent so as to maximize a long-term return received from an environment.
Recently, RL has been successfully investigated from single-agent problems to multi-agent tasks in a variety of fields, such as multi-player games~\cite{berner2019dota, ye2020towards}, sensor networks~\cite{zhang2011coordinated} and traffic light control~\cite{zhou2021drle}.
In this work, we focus on cooperative multi-agent reinforcement learning (MARL) with partial observability and communication constraints.
In such a setting, the agents are required to take action to interact with the environment in a decentralized manner. 
Generally, in MARL, the observed long-term return is characterized by stochasticity due to partial observations, changing policies of all the agents and environment model dynamics. 
Moreover, the stochasticity caused by actions will be intensified with the increasing number of agents. 
Due to the randomness in the long-term returns, it is preferable to model the value functions via distributions rather than the expectations.

To model the value functions via distributions, the current mainstream solution is distributional RL, which predicts the distribution over returns instead of a scalar mean value by leveraging either a categorical distribution~\cite{bellemare2017a} or a quantile function~\cite{dabney2018implicit}. 
Most of the studies in distributional RL focus on single-agent domains~\cite{bellemare2017a,dabney2018implicit,dabney2017distributional,yang2019fully}, which cannot be directly applied to the value-based MARL.
The reasons arise from two aspects:
(1) in value-based MARL, the individual distributional Q-values should be integrated into global distributional Q-value;
(2) the integration should guarantee the consistency between joint and individual greedy action selections in the global Q-value and individual Q-value, called \emph{Distributional-Individual-Global-Max} (DIGM) principle.

To our best knowledge, there exist few works focusing on distributional MARL~\cite{hu2020qr,qiu2021rmix}.
Only one recent work~\cite{sun2021dfac}, named DFAC, models both the individual and global Q-values from a distributional perspective, which decomposes the return distribution into the deterministic part (\emph{i.e.}, expected value) and stochastic part with mean zero.
To satisfy the DIGM principle, DFAC relies on a strong assumption that the expectation of global value can be fitted by the expectation of individual value, which does not necessarily hold in practice.
Taking a two-agent system as a toy example, the relationship between individual value and global value is $Z_{tot} = Relu(Z_1+Z_2)$.
\begin{equation*}
   \begin{aligned}
   \textbf{Case 1}: &\quad Z_{1}= \begin{cases}
    \begin{aligned}
       1 &\quad \text{prob}=0.5\\
       -1 &\quad \text{prob}=0.5
   \end{aligned} \end{cases}, \quad Z_{2}=1  \quad \rightarrow \quad \mathbb{E} \left[ Z_{tot1} \right] = 1\\
   \textbf{Case 2}: &\quad Z_{1}= \begin{cases}
    \begin{aligned}
       2 &\quad \text{prob}=0.5\\
       -2 &\quad \text{prob}=0.5
   \end{aligned} \end{cases}, \quad Z_{2}=1  \quad \rightarrow \quad \mathbb{E} \left[ Z_{tot2} \right] = 1.5
   \end{aligned}.
\end{equation*}
In the above two cases, the individual values follow different distributions with the same expectation, while the expectations of the global value are different. 
Thereby, it is difficult to fit the global value expectation only by the individual value expectations.

To this end, we propose a novel distributional MARL framework, \emph{i.e.}, parameterizing value function via \underline{M}ixture of \underline{C}ategorical distributions for MARL (abbreviated as MCMARL). 
Our method models both individual Q-value distributions and global Q-value distribution by categorical distribution.
In this way, the distributions of individual Q-values capture the uncertainty of the environment from each agent’s perspective while the distribution of global Q-value directly approximates the randomness of the total return.
To integrate the individual distributions into the global distribution, we define five basic operations, namely Weighting, Bias, Convolution, Projection and Function, which can realize the transformation of the distribution and the combination of multiple distributions.
These basic operations allow the generalization of expected value function factorization methods (\emph{e.g.}, VDN and QMIX) to their MCMARL variants without violating DIGM.

To evaluate the capability of MCMARL in distribution factorization, we first conduct a simple stochastic matrix game, where the true return distributions are known.
The results reveal that the distributions estimated by our method are very close to the true return distributions.
Beyond that, we perform experiments on a range of unit micromanagement benchmark tasks in StarCraft II~\cite{samvelyan19smac}.
The results on StarCraft II micromanagement benchmark tasks show that 
(1) our MCMARL framework is a more beneficial distributional MARL method than DFAC; 
(2) DQMIX (MCMARL variant of QMIX) always achieves the leading performance compared to the baselines. 
Furthermore, we analyze the impact of the hyperparameter---the size of the support set of categorical distribution, and figure out that a size of 51 is sufficient to obtain considerable performance.

\section{Background}
In this section, we introduce some background knowledge for convenience of understanding our method .
First, we discuss the problem formulation of a fully cooperative MARL task. 
Next, we introduce the concept of deep multi-agent Q-learning. 
Then, we present the CTDE paradigm and recent representative value function factorization methods in this field. 
Finally, we describe the concept of distributional RL and summarize the related studies.

\subsection{Decentralized Partially Observable Markov Decision Process}

We model a fully cooperative multi-agent task as a Decentralized Partially Observable Markov Decision Process (Dec-POMDP) \cite{oliehoek2016concise}, following the most recent works in cooperative MARL domain. 
Dec-POMDP can be described as a tuple $\mathcal{M}=\langle \mathcal{S}, \mathcal{Z}, \mathcal{A}, r, P, O, \mathcal{N},\gamma \rangle$, where $\mathcal{S}$ is a finite set of global states, $\mathcal{Z}$ is the set of individual observations and $\mathcal{A}$ is the set of individual actions. At each time step, each agent $i \in \mathcal{N} := \{g_1, \cdots, g_N\}$ selects an action $a_i \in \mathcal{A}$, forming a joint action $\boldsymbol{a}:=[a_i]_{i=1}^N \in \mathcal{A}^N$. This leads to a transition on the environment according to the state transition function $P(s^\prime|s,\boldsymbol{a}):\mathcal{S} \times \mathcal{A}^N \times \mathcal{S} \rightarrow [0, 1]$ and the environment returns a joint reward (\emph{i.e.}, team reward) $r(s,\boldsymbol{a}) : \mathcal{S} \times \mathcal{A} \rightarrow \mathbb{R}$ shared among all agents. $\gamma \in [0, 1)$ is the discount factor. 
Each agent $i$ can only receive an individual and partial observation $o_i \in \mathcal{Z}$, according to the observation function $O(s, i) :\mathcal{S} \times \mathcal{N} \rightarrow \mathcal{Z}$. And each agent $i$ has an action-observation history $\tau_i \in \mathcal{T} := (\mathcal{Z}\times\mathcal{A})^*$, on  which  it constructs its individual policy $\pi_i(a_i|\tau_i) :\mathcal{T}\times \mathcal{A} \rightarrow [0,1]$.
The objective of a fully cooperative multi-agent task is to learn a joint policy $\boldsymbol{\pi} := [\pi_i]_{i=1}^N$ so as to maximize the expected cumulative team reward.

\subsection{Deep Multi-Agent Q-Learning}

An early multi-agent Q-learning algorithm may be independent Q-learning (IQL) \cite{Tan1993MultiAgentRL}, which learns decentralized policy for each agent independently. IQL is simple to implement but suffers from non-stationarity of the environment and may lead to non-convergence of the policy. To this end, many multi-agent Q-learning algorithms \cite{sunehag2018value,rashid2018qmix,son2019qtran,yang2020qatten,wang2021qplex} are dedicated to learning a global Q-value function: 
\begin{equation}
\begin{aligned}
    Q_{tot}(\boldsymbol{\tau}, \boldsymbol{a})=\mathbb{E}_{s_{1:\infty}, \boldsymbol{a}_{1:\infty}} \left[ \sum_{t=0}^{\infty} \gamma^t r_t | s_0=s, \boldsymbol{a}_0 = \boldsymbol{a} \right],
\end{aligned}
\end{equation}
where $\boldsymbol{\tau}$ is the joint action-observation history and $r_t$ is the team reward at time $t$. 
Similar to DQN \cite{mnih2015human}, deep multi-agent Q-learning algorithms represent the global Q-value function with a deep neural network parameterized by $\theta$, and then use a replay memory to store the transition tuple $(\boldsymbol{\tau}, \boldsymbol{a}, r, \boldsymbol{\tau}^{\prime})$. Parameters $\theta$ are learnt by sampling a batch of transitions $\{(\boldsymbol{\tau}^{(i)}, \boldsymbol{a}^{(i)}, r^{(i)}, \boldsymbol{\tau}^{\prime (i)})\}_{i=1}^n$ to minimize the following TD error:
\begin{equation}
\begin{aligned}
    \mathcal{L}(\theta) = \sum_{i=1}^n \left[  \left( y_{tot}^{(i)} - Q_{tot}\left(\boldsymbol{\tau}^{(i)}, \boldsymbol{a}^{(i)};\theta\right)\right)^2  \right],
\end{aligned}
\end{equation}
where $y_{tot}^{(i)} = r^{(i)}+\gamma \max_{\boldsymbol{a}^{\prime}} Q_{tot}\left(\boldsymbol{\tau}^{\prime (i)}, \boldsymbol{a}^{\prime}; \theta^-\right)$. 
$\theta^-$ represent the parameters of target network that are copied every $C$ steps from $\theta$.
The joint policy can be derived as: $\boldsymbol{\pi}(\boldsymbol{\tau}) = \arg\max_{\boldsymbol{a}} Q_{tot}(\boldsymbol{\tau}, \boldsymbol{a}; \theta)$. 

\subsection{CTDE and Value Function Factorization}
In cooperative MARL, fully decentralized methods \cite{Tan1993MultiAgentRL,tampuu2017multiagent} are scalable but suffer from non-stationarity issue. 
On the contrary, fully centralized methods \cite{guestrin2002coordinated,kok2006collaborative} mitigate the non-stationarity issue but encounter the challenge of scalability, as the joint state-action space grows exponentially with the number of agents. 
To combine the best of both worlds, a popular paradigm called centralized training with decentralized execution (CTDE) has drawn substantial attention recently. 
In CTDE, agents take actions based on their own local observations and are trained to coordinate their actions in a centralized way.
During execution, the policy of each agent only relies on its local action-observation history, which guarantees the decentralization.

Recent value-based MARL methods realize CTDE mainly by factorizing the global Q-value function into individual Q-value functions \cite{sunehag2018value,rashid2018qmix,son2019qtran}. 
To ensure the collection of individual optimal actions of each agent during execution is equivalent to the optimal actions selected from global Q-value, value function factorization methods have to satisfy the following IGM \cite{son2019qtran} condition:
\begin{equation}
\begin{aligned}
    \arg\max_{\boldsymbol{a}} Q_{tot}(\boldsymbol{\tau}, \boldsymbol{a}) = \left(
    \begin{matrix}
    \arg\max_{a_1} Q_1(\tau_1, a_1) \\
    \vdots \\
    \arg\max_{a_N} Q_N(\tau_N, a_N)
    \end{matrix}
    \right).
    \label{igm}
\end{aligned}
\end{equation}

As the first attempt of this stream, VDN~\cite{sunehag2018value} represents the global Q-value function as a sum of individual Q-value functions.
Considering that VDN ignores the global information during training, QMIX~\cite{rashid2018qmix} assigns the non-negative weights to individual Q-values with a non-linear function of the global state.
These two factorization methods are sufficient to satisfy Eq. (\ref{igm}) but inevitably limit the global Q-value function class they can represent due to their structural constraint. 
To address the representation limitation, QTRAN proposes to learn a state-value function and transform the original global Q-value function $Q_{tot}$ into an easily decomposable one $Q_{tot}^{\prime}$ that shares the same optimal actions with $Q_{tot}$ \cite{son2019qtran}. 
However, the computationally intractable constraint imposed by QTRAN may lead to poor performance in complex multi-agent tasks.

\subsection{Distributional RL}
Distributional RL aims to approximate the distribution of returns (\emph{i.e.}, the discounted cumulative rewards) denoted by a random variable $Z(s, a)$, whose expectation is the scalar value function $Q(s, a)$. 
Similar to the Bellman equation of Q-value function, the distributional Bellman equation can be defined by
\begin{equation}
\begin{aligned}
    Z(s,a) \overset{D}{:=} R(s, a) + \gamma Z(s^\prime, a^\prime),
    \label{bellman}
\end{aligned}
\end{equation}
where $s^\prime \sim P(\cdot\,|\,s,\, a),\, a^\prime \sim \pi(\cdot\,|\,s^\prime)$, and $A \overset{D}{=} B$ denotes that random variables $A$ and $B$ have the same distribution. 
As revealed in Eq. (\ref{bellman}), $ Z(s,a)$ involves three sources of randomness : the reward $R(s, a)$, the transition $ P(\cdot\,|\,s,\, a)$, and  the next-state value distribution $Z(s^\prime, a^\prime)$ \cite{bellemare2017a}.
Then, we have the distributional Bellman optimality operator $T^*$ as follows:
\begin{equation}
\begin{aligned}
    T^*Z(s,a) \overset{D}{:=} R(s, a) + \gamma Z(s^\prime, \arg\max_{a^\prime} \mathbb{E}\left[Z(s^\prime, a^\prime)\right]).
    \label{opti_bellman}
\end{aligned}
\end{equation}

Based on the distributional Bellman optimality operator, the objective of distributional RL is to reduce the distance between the distribution $ Z(s,a)$ and the target distribution $T^*Z(s,a)$. 
Therefore, a distributional RL algorithm must address two issues: how to parameterize the return distribution and how to choose an appropriate metric to measure the distance between two distributions. 
To model the return distribution, many RL methods in SARL domain are proposed with promising results \cite{bellemare2017a,dabney2018implicit,dabney2017distributional,yang2019fully}.  
In this paper, we employ the categorical distribution \cite{bellemare2017a}, which represents the distribution with probability masses placed on a discrete set of possible returns, and then minimize the Kullback–Leibler (KL) divergence between the Bellman target and the current estimated return distribution.

\section{Method}
In this section, we first define five basic operations on the distribution of random variables, that satisfy the DIGM principle.
Based on the operations, we give an introduction to our MCMARL framework and illustrate the variants of VDN and QMIX under MCMARL framework.
Furthermore, we briefly present the training and execution strategy. 



\subsection{Basic Operations on Distribution}
Let $X$ be a discrete random variable, following a categorical distribution, denoted by $X \sim (M,V_{min},V_{max},\mathbf{P})$, where $M \in \mathbb{N}$, $V_{min},V_{max}\in \mathbb{R}$.
The support of $X$ is a set of atoms $ \{ x_j = V_{min} + j \triangle x : 0 \leq j < M \}$, where $\triangle x \coloneqq \frac{V_{max}-V_{min}}{M-1}$ and $\mathbf{P}$ is the atom probability, \emph{i.e.,}
\begin{equation}
\begin{aligned}
    X = x_{j} \quad  w.p. \quad  p_j.
\end{aligned}
\label{ali:Z}
\end{equation}

\begin{figure*}[t]

	\centering
	\subfigure[]{
		\includegraphics[width=0.3\textwidth]{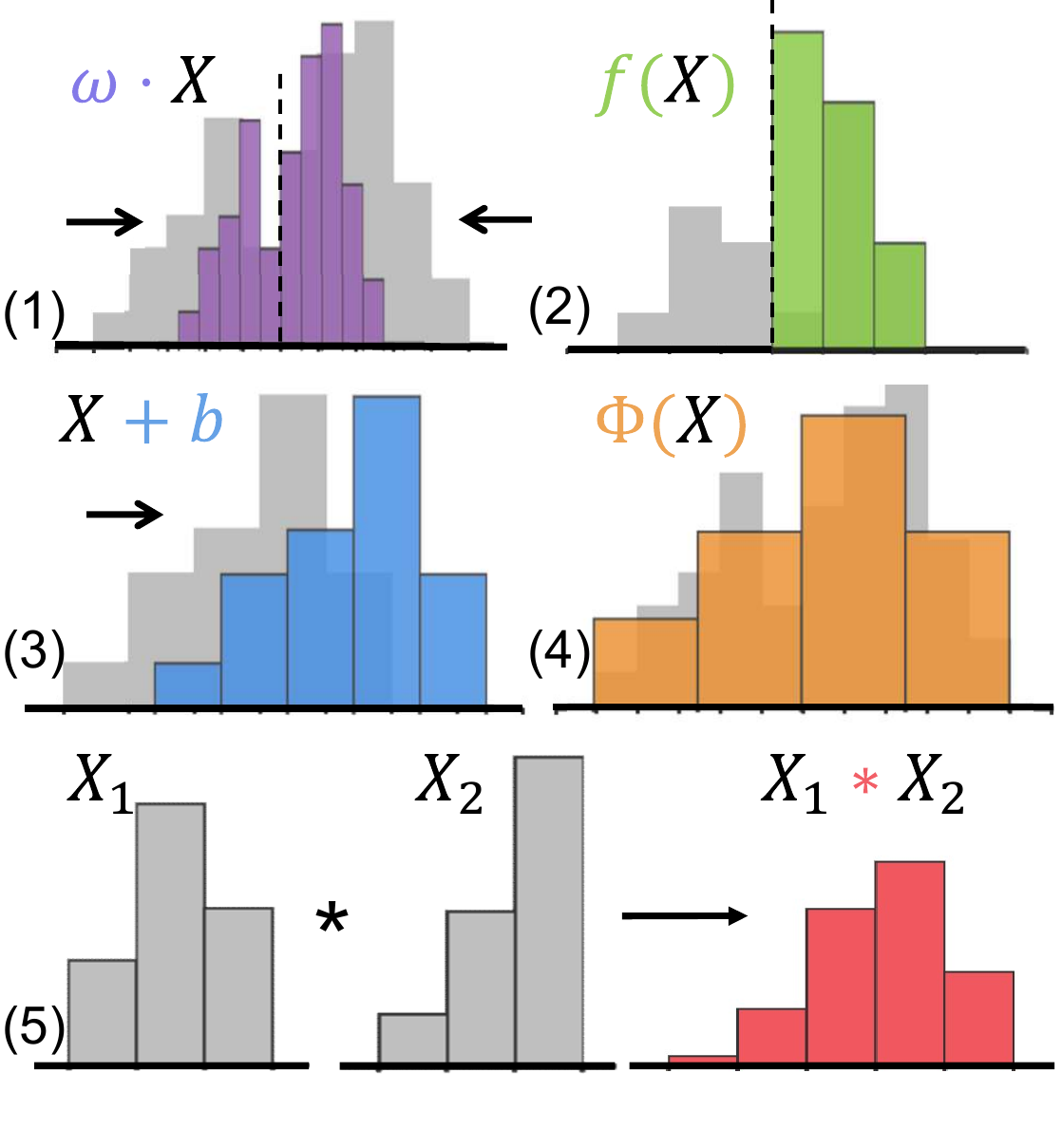}
		\label{fig:op}
	}
	\subfigure[]{
		\includegraphics[width=0.65\textwidth]{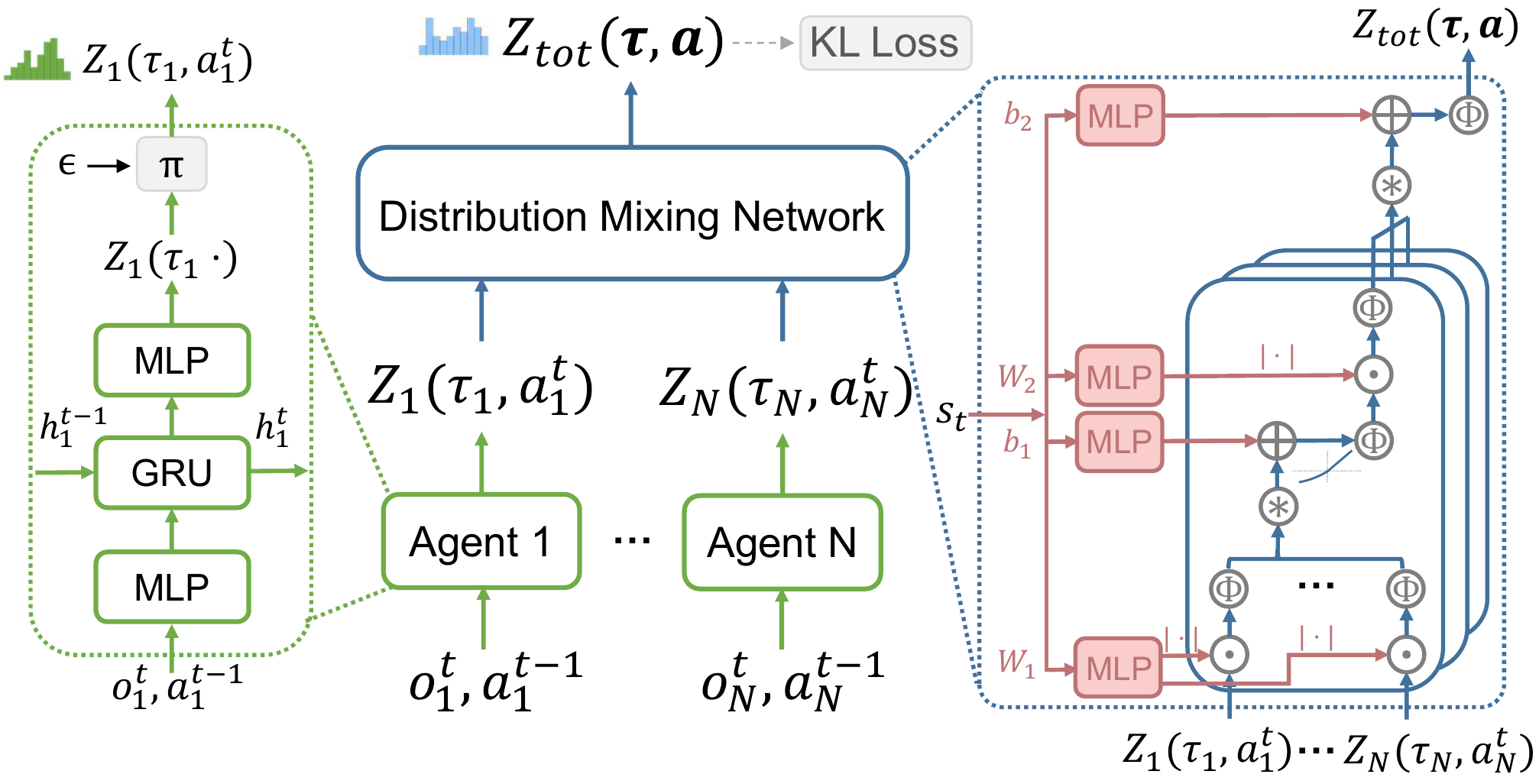}
		\label{fig:struc}
	}
	\caption{(a) The five basic operations on distribution of random variables. Operations (1)-(5) represent Weighting, Function, Bias, Projection and Convolution, respectively. (b) The overall DQMIX architecture, an example of MCMARL variant of the value-based factorization model.}
\end{figure*}

To apply transformation and combination to random variables with categorical distribution, we define five basic operations as illustrated in Figure~\ref{fig:op}.

\noindent\textbf{Operation 1. [Weighting]}  
Analogous to the scaling operation to a scalar variable, the weighting operation $W_w$ to scale up a discrete random variable by $w \in \mathbb{R}$ is defined as follows:
\begin{equation}
    W_{w}X \coloneqq w x_j \quad w.p.\quad  p_j.
\end{equation}
The Weighting operation over distribution can be abbreviated as $w \cdot X$.

\noindent\textbf{Operation 2. [Bias]}  
Analogous to the panning operation to a scalar variable, the bias operation $B_{b}$ to pan a discrete random variable by $b \in \mathbb{R}$ is defined as follows:
\begin{equation}
    B_{b}X \coloneqq  x_j+b \quad w.p.\quad  p_j,
\end{equation}
which is abbreviated as $ X+b$.

\noindent\textbf{Operation 3. [Convolution]}  
To combine the two random variables $X_{1} \sim (M_{1},V_{1,min},V_{1,max},\mathbf{P}_{1})$ and $X_{2} \sim (M_{2},V_{2,min},V_{2,max},\mathbf{P}_{2})$ with the same atom interval $\triangle x$, we define the convolution operation $Conv(\cdot,\cdot)$ as follows:
\begin{equation}
\begin{aligned}
& Conv(X_{1},X_{2}) \coloneqq  x^*_j \quad w.p.\quad  p^*_j,\\
& x^*_j = V_{1,min} +V_{2,min} + \triangle x\cdot j,\\
\end{aligned}
\end{equation}
where $0 \leq j < M_{1}+M_{2}-1$.
Let $\hat{M}  \coloneqq M_{1}+M_{2}-1$.
If $M_{1} \ge M_{2}$, then
\begin{equation}
p^*_j = \begin{cases}
\sum_{k=0}^j p_{1,k} p_{2,j-k} & 0\le j <M_{2}\\
\sum_{k=j-M_{2}+1}^{j} p_{1,k} p_{2,j-k} & M_{2}\le j <M_{1}\\
\sum_{k=j-M_{2}+1}^{M_{1}-1} p_{1,k} p_{2,j-k} & M_{1}\le j <\hat{M}\\
\end{cases}.
\end{equation}
If $M_{1} < M_{2}$, then
\begin{equation}
p^*_j = \begin{cases}
\sum_{k=0}^j p_{1,j-k} p_{2,j} & 0\le j <M_{1}\\
\sum_{k=j-M_{1}+1}^{j} p_{1,j-k} p_{2,j} & M_{1}\le j <M_{2}\\
\sum_{k=j-M_{1}+1}^{M_{2}-1} p_{1,j-k} p_{2,j} & M_{2}\le j <\hat{M}\\
\end{cases}.
\end{equation}
$Conv(X_{1},X_{2})$ is abbreviated as $ X_{1}*X_{2}$.

\noindent\textbf{Operation 4. [Projection]} 
To map the random variable distribution of $[x_j]$ to atoms $[\hat{x}_k]$ where $ \{ \hat{x}_k = \hat{V}_{min} + k \triangle \hat{x} : 0 \leq k < K \}$, $\triangle \hat{x} \coloneqq \frac{\hat{V}_{max}-\hat{V}_{min}}{K-1}$, we define the projection operation $\Phi_{[\hat{x}_k]}$ as follows :
\begin{equation}
\Phi_{[\hat{x}_k]} X  \coloneqq  \hat{x}_k \quad w.p.\quad  
\sum_j\left[ 1- \frac{|[x_j]^{\hat{V}_{max}}_{\hat{V}_{min}}-\hat{x}_k|}{\triangle \hat{x}} \right]^{1}_{0} p_j,
\end{equation}
where $[\cdot]^a_b$ bounds its argument in the range $[a,b]$. 

\noindent\textbf{Operation 5. [Function]}  
To apply non-linear operation over a random variable, we define the function operation $F_{f}, f:\mathbb{R} \rightarrow \mathbb{R}$ as follows:
\begin{equation}
    F_{f} X \coloneqq  f(x_j) \quad w.p.\quad  p_j,
\end{equation}
which is abbreviated as $f(X)$.

Furthermore, we give theoretical proof that the five basic operations satisfy the DIGM principle.
Correspondingly, the network structure composed of these five basic operations also satisfies the DIGM principle.
Due to the page limit, the detailed proof is attached in Appendix~\ref{appendix1}.

\subsection{Framework of MCMARL}

For each agent, we parameterize its individual Q-value distribution with categorical distribution, which has high flexibility to approximate any shape of the distribution.
Assume the support set of the agent's distribution, denoted as $[z]=\{z_{1}, z_{2}, \cdots, z_{M} | z_{1} \leq \cdots \leq z_{M}, M \in \mathbb{N}\}$, is uniformly distributed over the predefined range $[V_{min}, V_{max}]$, where $V_{min}, V_{max} \in \mathbb{R}$ are the minimum and maximum returns, respectively. 
Note that all the individual Q-value distributions share the same support set.

Based on these assumptions, learning individual distributions is equivalent to learning the atom probabilities.
For each agent, there is one agent network $\theta : \mathcal{Z} \times \mathcal{A} \rightarrow \mathbb{R}^M$, that estimates the probabilities of atoms in the support set.
Taking agent $i$ as an example, at each time step, its agent network receives the current individual observation $o_i^t$ and the last action $a_i^{t-1}$ as input, and generates the atom probabilities as follows:
\begin{equation}
    Z_{i}(\tau_{i},a_{i}) = z_{j} \quad  w.p. \quad p_{i,j}(\tau_{i},a_{i}) \coloneqq \frac{e^{\theta_j(\tau_{i},a_{i})}}{\sum_k{e^{\theta_k(\tau_{i},a_{i})}}}, 
\end{equation}
where $j \in \{1, \cdots, M\}$ and $\tau_i$ denotes the agent $i$'s action-observation history.
For scalability, parameters $\theta$ are shared among all agents. 

\begin{figure*}[t]
	\centering
	\includegraphics[width=0.98\linewidth]{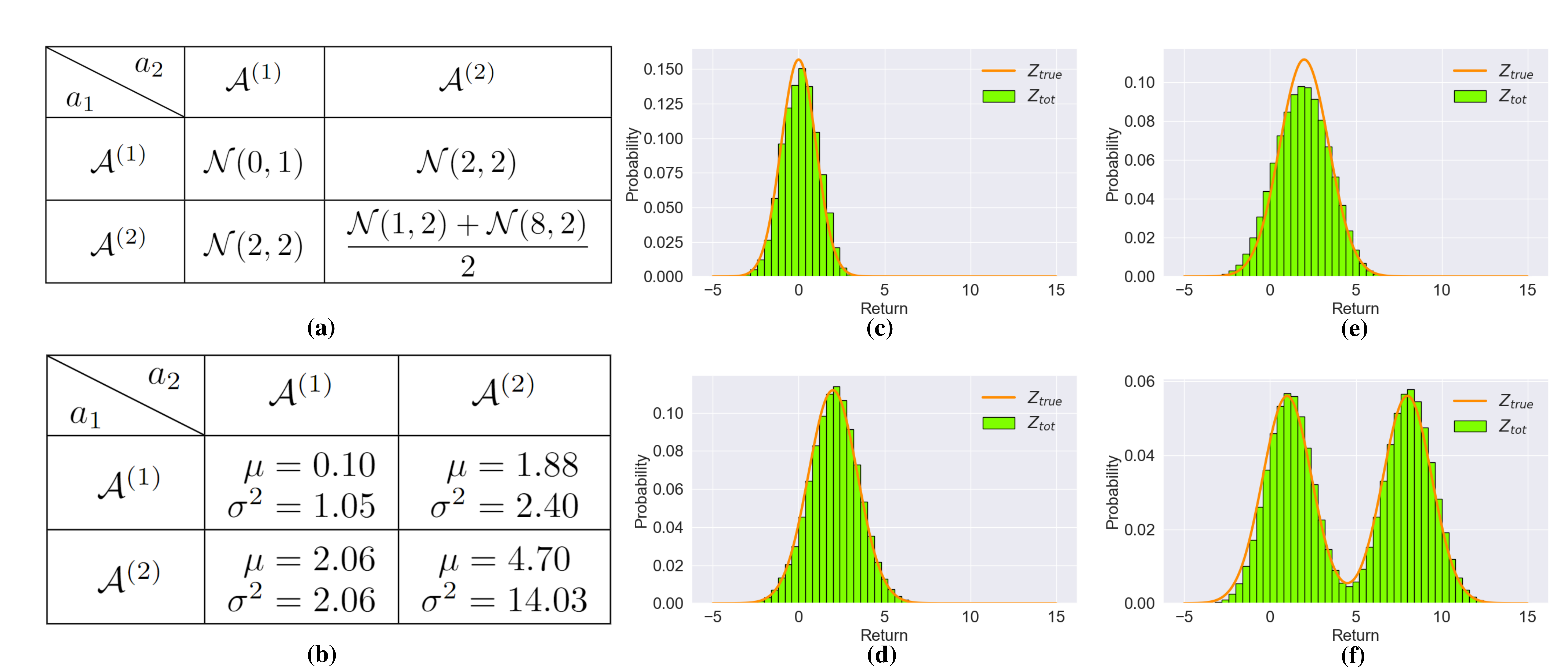}
	\caption{(a) The stochastic matrix game. Each agent $i \in \{ 1, 2 \}$ takes an action $a_i \in \{ \mathcal{A}^{(1)}, \mathcal{A}^{(2)}\}$ and then receives a joint reward that is sampled from the corresponding distribution in the matrix. $\mathcal{N}(\mu, \sigma^2)$ represents the gaussian distribution with mean $\mu$ and variance $\sigma^2$. The joint action $\langle \mathcal{A}^{(2)}, \mathcal{A}^{(2)} \rangle$ results in a  bimodal distribution $\frac{\mathcal{N}(1,2)+\mathcal{N}(8,2)}{2}$, which means that the joint reward is sampled from $\mathcal{N}(1,2)$ or $\mathcal{N}(8,2)$ with equal probability. (b) The learned global Q-value distribution of DQMIX. $\mu$ and $\sigma^2$ indicate the sampled mean and the sampled variance, respectively. (c)-(f) The total return distributions of joint action $\langle \mathcal{A}^{(1)}, \mathcal{A}^{(1)} \rangle$, $\langle \mathcal{A}^{(2)}, \mathcal{A}^{(1)} \rangle$, $\langle \mathcal{A}^{(1)}, \mathcal{A}^{(2)} \rangle$ and $\langle \mathcal{A}^{(2)}, \mathcal{A}^{(2)} \rangle$, respectively. The orange line represents the true return distribution (\emph{i.e.}, the sampled joint reward distribution). The green histogram shows the global Q-value distribution learned by DQMIX.}
	\label{matrix}
\end{figure*}

Given the five basic operations, the variants of the existing value-based models under our MCMARL framework can be designed based the basic operations over the individual distributions, as shown in Figure~\ref{matrix}(b).
Here we introduce DVDN and DQMIX, which are the corresponding MCMARL variants of VDN and QMIX respectively.

VDN sums up the individual values as the global value and its MCMARL variant, DVDN, is to apply the Convolution operation over the individual value distributions followed by the Projection operation.
\newcommand{\shorteqnote}[1]{ & & & & & & & & & & & \text{\small\llap{#1}}}
\begin{equation}
\begin{aligned}
    1. \ A_{tot} & := Z_{1} * Z_{2} * \cdots * Z_{N},\shorteqnote{(\text{Convolution})}\\
    2. \ Z_{tot} & := \Phi_{[z]}(A_{tot}).  \shorteqnote{(\text{Projection})}
    \nonumber
\end{aligned}
\end{equation}
Corresponding to the design of VDN, DVDN simply takes a sum of the individual randomness as the global randomness, which relies on the fact that the individual Q-values are independent.
However, the independence of individual Q-value distributions does not necessarily hold, which might limit the performance of DVDN (see A.2 for details).

DQMIX, the MCMARL variant of QMIX, mixes the individual Q-values distributions into global Q-value distribution $Z_{tot}$ by leveraging a multi-layer neural network.
The sequence of operations of $k^{th}$ layer is formulated as follows:
\begin{equation}
\begin{aligned}
    & 1. \ A_{i,j}^k := w_{i,j}^k \cdot Z_i^k, \shorteqnote{(\text{Weighting})} \\
    & 2. \ B_{i,j}^k := \Phi_{[z]}(A_{i,j}^k), \shorteqnote{(\text{Projection})}\\
    & 3. \ C_j^k := B_{1,j}^k * B_{2,j}^k * \cdots * B_{N_k,j}^k,\shorteqnote{(\text{Convolution})}\\
    & 4. \ D_j^k := C_j^k + b^k_j, \shorteqnote{(\text{Bias})}\\
    & 5. \ E_j^k := f\left(D_j^k\right),  \shorteqnote{(\text{Function})}\\
    & 6. \ Z_j^{k+1} := \Phi_{[z]}(E_j^k),  \shorteqnote{(\text{Projection})}
    \nonumber
\end{aligned}
\end{equation}
where $i \in \{1, \cdots, N_k\}, j \in \{1, \cdots, N_{k+1}\}$, and $N_k$ is the number of input distributions of $k^{th}$ layer.
The parameters $w_{i,j}^k \in \mathbb{R}$ and $b^k_j \in \mathbb{R}$ are generated by the $k^{th}$ hypernetwork conditioned on the global state.
Note that each hypernetwork that generates $w_{i,j}^k$ is followed by an absolute activation function, which guarantees that the parameters of the Weighting operations are non-negative.

\subsection{Training and Execution}

In the training phase, each agent $i$ interacts with the environment using the $\epsilon$-greedy policy over the expectation of individual Q-value distribution, \emph{i.e.}, $\mathbb{E}[Z_i] = \sum_j p_{i,j} z_j$. 
The transition tuple $(\boldsymbol{\tau}, \boldsymbol{a}, r, \boldsymbol{\tau}^{\prime})$ is stored into a replay memory. 
Then, the learner randomly fetches a batch of samples from the replay memory.
The network is optimized by minimizing the sample loss, \emph{i.e.}, the cross-entropy term of KL divergence:
\begin{equation}
\begin{aligned}
    D_{\text{KL}}(\Phi T^* Z_{tot}(\boldsymbol{\tau}, \boldsymbol{a}; \theta^-) \,\,||\,\, Z_{tot}(\boldsymbol{\tau}, \boldsymbol{a}; \theta)),
\end{aligned}
\end{equation}
where $T^* Z_{tot}(\boldsymbol{\tau}, \boldsymbol{a}; \theta^-)$ is the Bellman target according to Eq. (\ref{opti_bellman}), $\theta^-$ are the parameters of a target network that are periodically copied from $\theta$, and $\Phi$ is the projection of Bellman target onto the support of $Z_{tot}(\boldsymbol{\tau}, \boldsymbol{a}; \theta)$.

Since our method satisfies the DIGM condition, the policy learned during centralized training can be directly applied to execution.
During the execution phase, each agent chooses a greedy action $a_i$ at each time step with respect to $\mathbb{E}[Z_i]$. 


\section{Experiments}

In this section, we first present our method on a simple stochastic matrix game to show MCMARL’s ability to approximate the true return distribution and the benefits of modeling the value distribution. 
Then, we further evaluate the efficacy of MCMARL on StarCraft Multi-Agent Challenge (SMAC) benchmark environment \cite{samvelyan19smac}.
Finally, we study the impact of atom number on our approach. 
All of our experiments are conducted on GeForce RTX 2080Ti GPU.
The implementation code is available in the supplementary material.

\begin{figure*}[t]
	\centering
	\subfigure[corridor]{
		\includegraphics[width=0.28\textwidth]{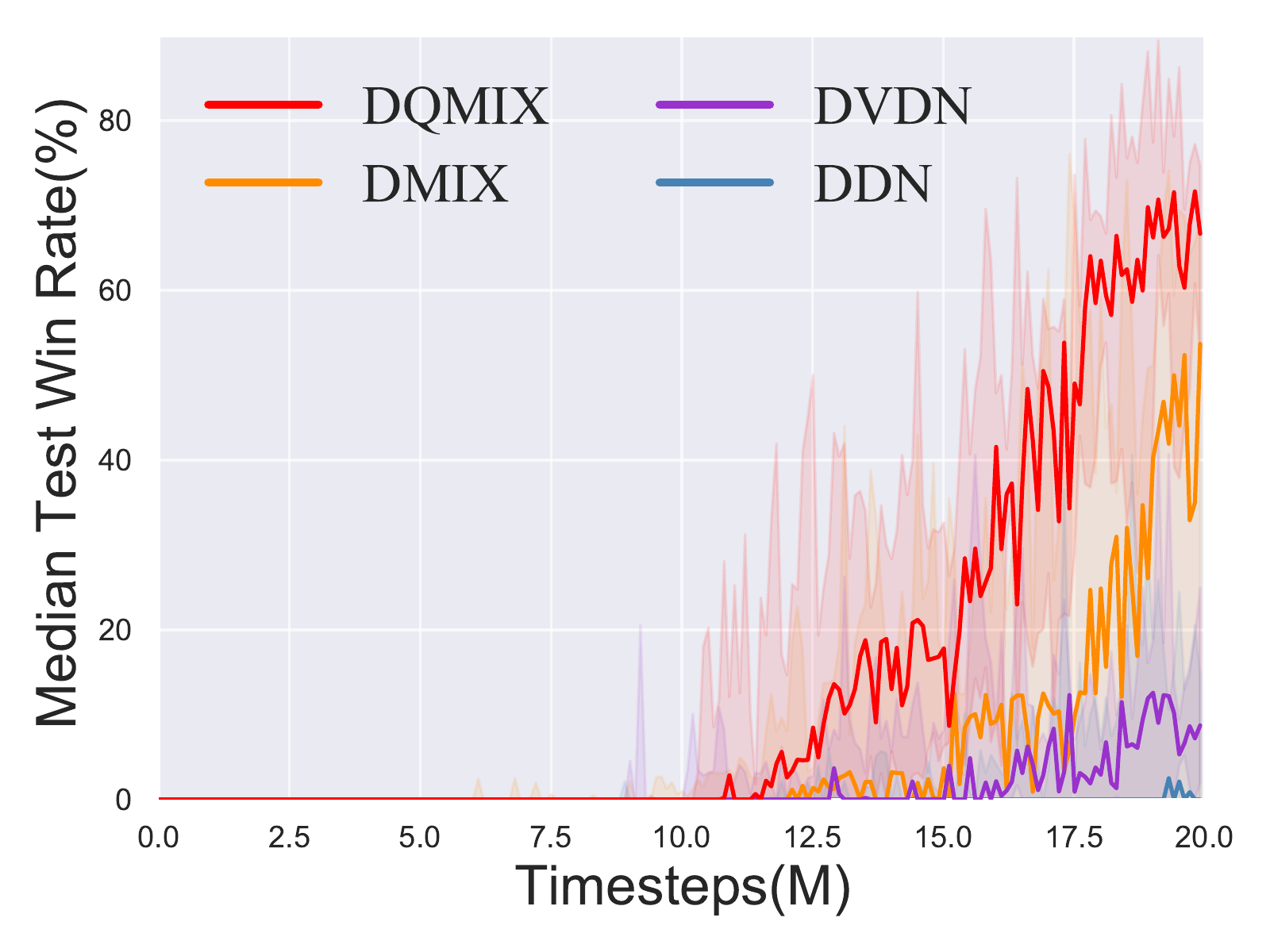}
		\label{fig:corridor}
	}
	\subfigure[10m\_vs\_11m]{
		\includegraphics[width=0.28\textwidth]{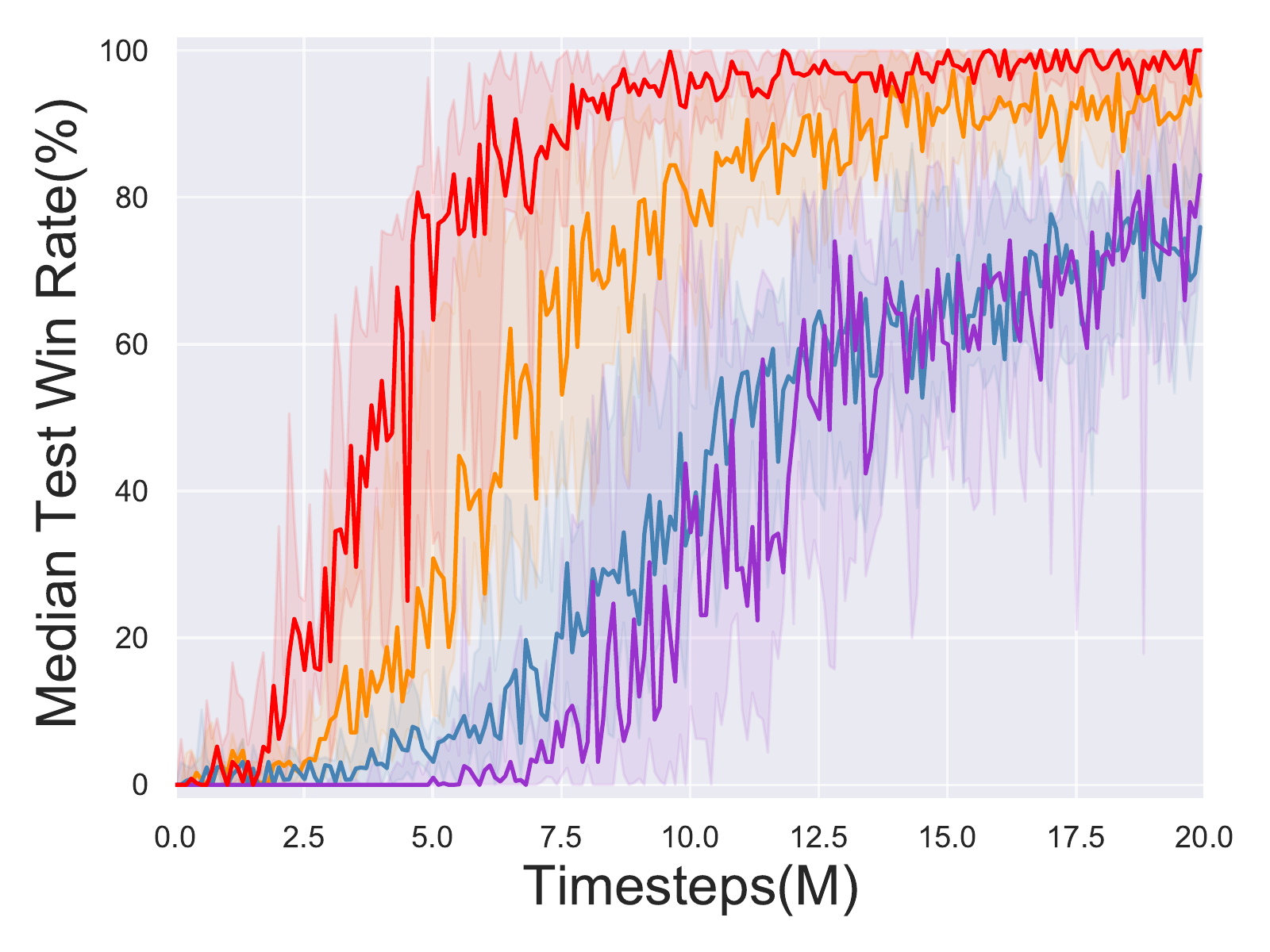}
		\label{fig:10m_vs_11m}
	}
	\subfigure[5m\_vs\_6m]{
		\includegraphics[width=0.28\textwidth]{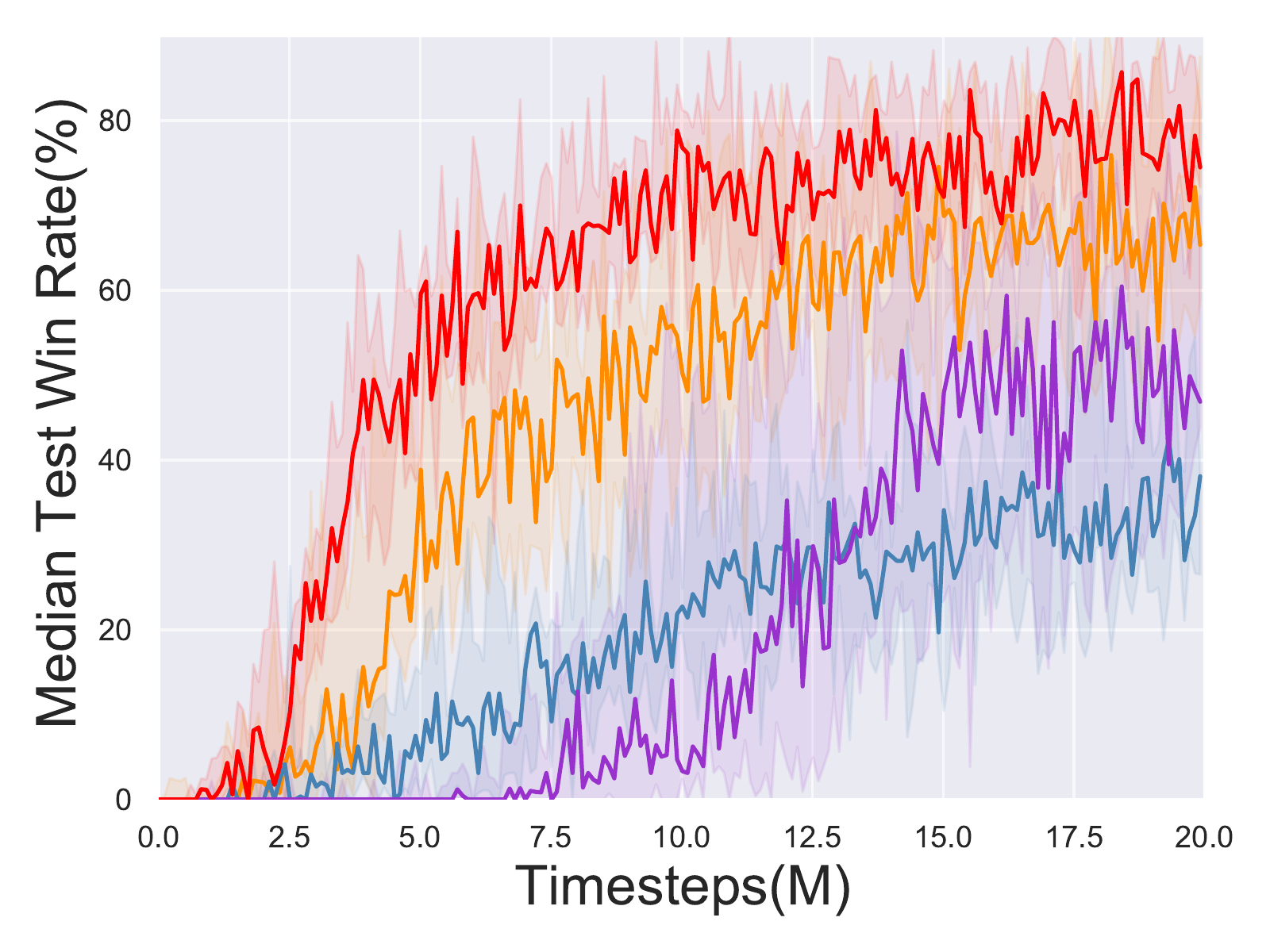}
		\label{fig:5m_vs_6m}
	}
	\subfigure[2s\_vs\_1sc]{
		\includegraphics[width=0.28\textwidth]{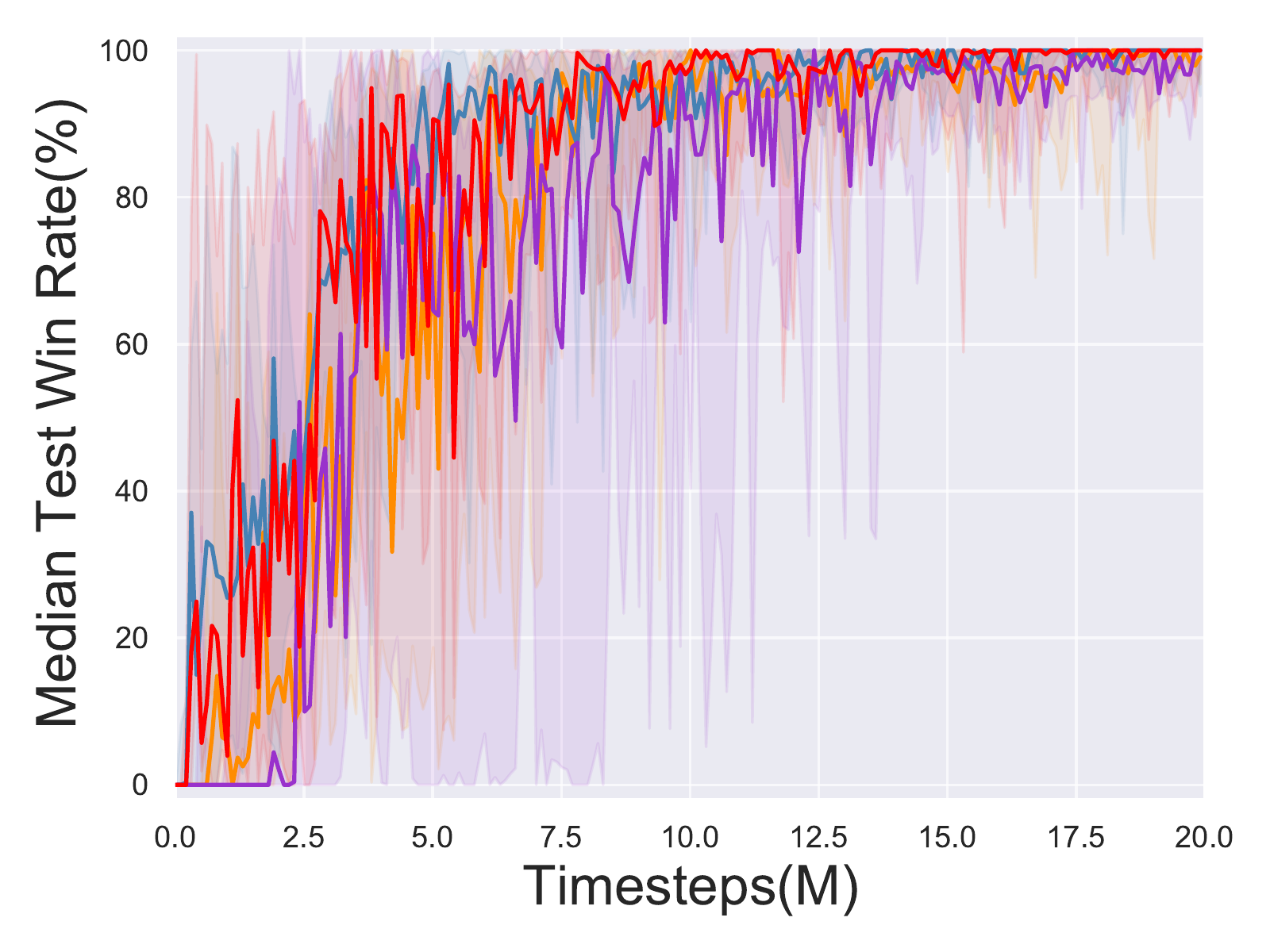}
		\label{fig:2s_vs_1sc}
	}
	\subfigure[2c\_vs\_64zg]{
		\includegraphics[width=0.28\textwidth]{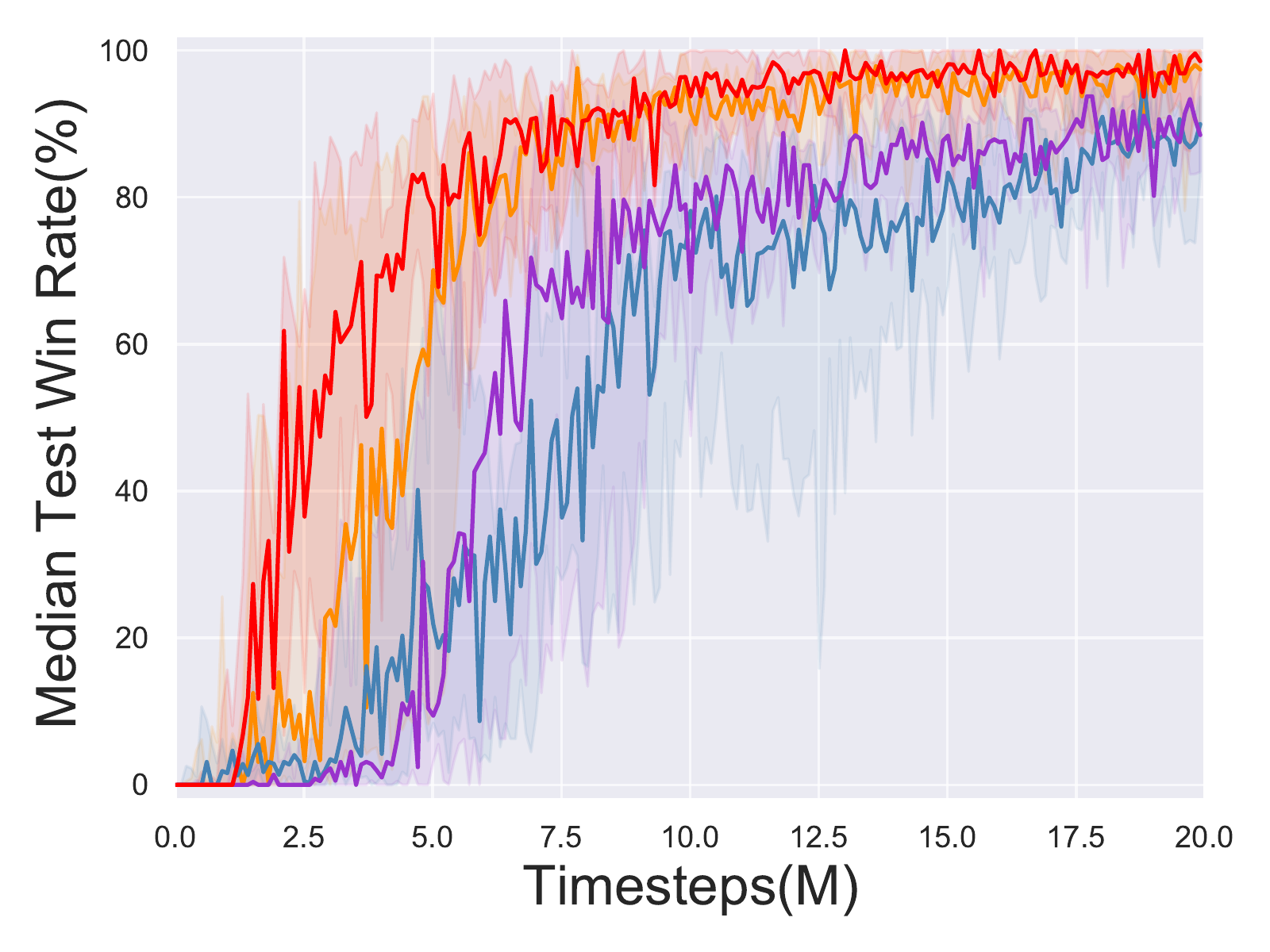}
		\label{fig:2c_vs_64zg}
	}
	\subfigure[2s3z]{
		\includegraphics[width=0.28\textwidth]{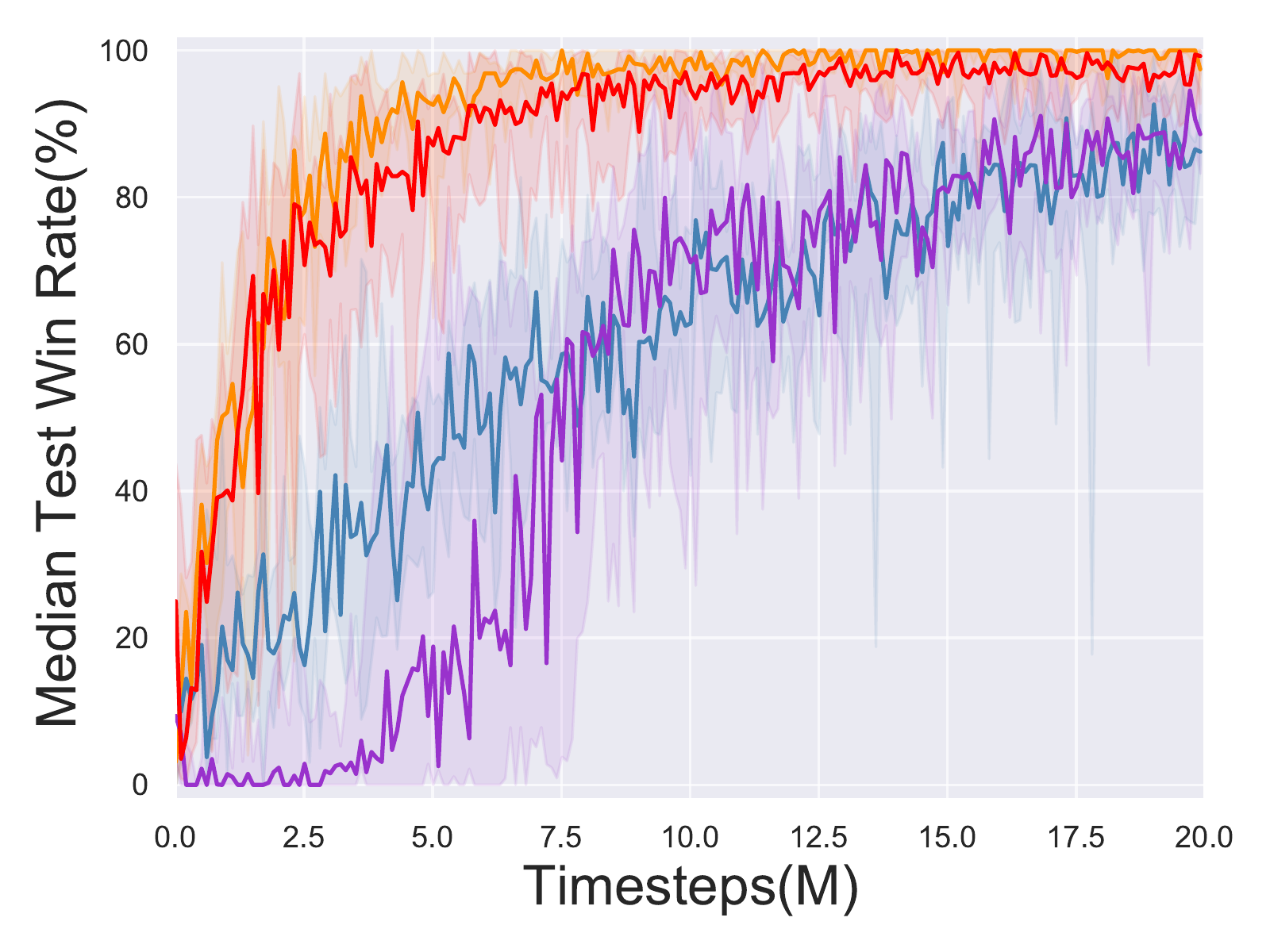}
		\label{fig:2s3z}
	}
	\setlength{\belowcaptionskip}{-0.5cm}
	\caption{The test win rate curves on the SMAC benchmark. The solid line shows the median win rate and the shadow area represents the min to max win rate on 5 random seeds.}
	\label{fig:smac}
\end{figure*}

\subsection{Evaluation on Stochastic Matrix Game}

Matrix game is widely adopted to test the effectiveness of the methods~\cite{sun2021dfac,rashid2018qmix,son2019qtran}.
To demonstrate the capacity of MCMARL to approximate the true return distribution, we design a two-agent stochastic matrix game.
Specifically, two agents jointly take actions and will receive a joint reward, which follows a distribution rather than a deterministic value.
Here, we set the joint reward to follow a normal distribution or a mixture of normal distributions, as illustrated in Figure~\ref{matrix}(a).

Take the MCMARL variant of QMIX as an example, we train DQMIX on the matrix game for 2 million steps with full exploration (\emph{i.e.}, $\epsilon$-greedy exploration with $\epsilon=1$).
Full exploration ensures that DQMIX can explore all available game states, such that the representational capacity of the state-action value distribution approximation remains the only limitation~\cite{rashid2018qmix}.
As shown in Figure~\ref{matrix}(b), the learned global Q-value distributions are close to the true return distributions in terms of the mean and variance. 
Moreover, we visualize the true return distribution and the learned distribution of joint action in Figure~\ref{matrix}(c)-(f).
It can be observed that the estimated distributions are extremely close to the true ones, which can not be achievable by expected value function factorization methods.

\subsection{Evaluation on SMAC Benchmark}

\begin{table*}[!ht]
        \centering
        \resizebox{\linewidth}{!}{
        \renewcommand{\arraystretch}{1.25}
        \begin{tabular}{|cc|cccccc|}
        \hline
        \multicolumn{2}{|c|}{\multirow{2}{*}{Method}}               &  \multicolumn{6}{c|}{Scenario}                                                                                                      \\ \cline{3-8} 
        \multicolumn{2}{|c|}{}                                &                                 \multicolumn{1}{c|}{corridor} & \multicolumn{1}{c|}{10m\_vs\_11m} & \multicolumn{1}{c|}{5m\_vs\_6m} & \multicolumn{1}{c|}{2s\_vs\_1sc} & \multicolumn{1}{c|}{2c\_vs\_64zg} & 2s3z \\ \hline
        \multicolumn{1}{|c|}{\multirow{2}{*}{DFAC \cite{sun2021dfac}}}   & DDN                               & \multicolumn{1}{c|}{3.68$\pm$5.67}         & \multicolumn{1}{c|}{77.02$\pm$4.19}   & \multicolumn{1}{c|}{34.95$\pm$4.69}           & \multicolumn{1}{c|}{98.75$\pm$2.50}     & \multicolumn{1}{c|}{88.63$\pm$4.41}     &  \multicolumn{1}{c|}{87.06$\pm$2.40}      \\ \cline{2-8} 
        \multicolumn{1}{|c|}{}                        & DMIX                             & \multicolumn{1}{c|}{35.49$\pm$29.38}         & \multicolumn{1}{c|}{93.14$\pm$7.69}   & \multicolumn{1}{c|}{68.48$\pm$10.25}           & \multicolumn{1}{c|}{98.39$\pm$1.61}     & \multicolumn{1}{c|}{96.37$\pm$3.91}             &  \multicolumn{1}{c|}{96.71$\pm$3.13}      \\ \hline
        \multicolumn{1}{|l|}{\multirow{2}{*}{MCMARL}} & DVDN                               & \multicolumn{1}{c|}{12.79$\pm$9.40}         & \multicolumn{1}{c|}{84.79$\pm$3.91}  & \multicolumn{1}{c|}{52.38$\pm$10.43}           & \multicolumn{1}{c|}{99.38$\pm$1.25}     & \multicolumn{1}{c|}{89.57$\pm$5.13}     & \multicolumn{1}{c|}{88.86$\pm$4.04}       \\ \cline{2-8} 
        \multicolumn{1}{|l|}{}                        & DQMIX                          & \multicolumn{1}{c|}{64.95$\pm$8.35}         & \multicolumn{1}{c|}{98.35$\pm$2.13}   & \multicolumn{1}{c|}{75.59$\pm$3.45}           & \multicolumn{1}{c|}{99.88$\pm$0.23}     & \multicolumn{1}{c|}{98.67$\pm$1.22}             &   \multicolumn{1}{c|}{99.07$\pm$1.20}     \\ \hline
        \end{tabular}
        }
        \caption{The average values and standard deviations of test win rate (\%) of 5 independent runs after 20 million training timesteps of algorithms under different distributional frameworks.}
        \label{tab:smac_dis}
\end{table*}

\begin{table*}[!ht]
        \centering
        \resizebox{\linewidth}{!}{
        \renewcommand{\arraystretch}{1.25}
        \begin{tabular}{|c|cccccc|}
        \hline
        \multicolumn{1}{|c|}{\multirow{2}{*}{Method}}               &  \multicolumn{6}{c|}{Scenario}                                                                                                    
        \\ \cline{2-7} 
        \multicolumn{1}{|c|}{}                                &                                  \multicolumn{1}{c|}{corridor} & \multicolumn{1}{c|}{10m\_vs\_11m} & \multicolumn{1}{c|}{5m\_vs\_6m} & \multicolumn{1}{c|}{2s\_vs\_1sc} & \multicolumn{1}{c|}{2c\_vs\_64zg} & \multicolumn{1}{c|}{2s3z} \\ \hline
        \multicolumn{1}{|c|}{IQL \cite{Tan1993MultiAgentRL}}                                                           & \multicolumn{1}{c|}{12.57$\pm$12.14}         & \multicolumn{1}{c|}{72.19$\pm$6.25}   & \multicolumn{1}{c|}{50.96$\pm$7.97}           & \multicolumn{1}{c|}{99.42$\pm$0.79}     & \multicolumn{1}{c|}{84.81$\pm$3.72}             &   \multicolumn{1}{c|}{81.69$\pm$3.69}     \\ \hline
        \multicolumn{1}{|c|}{VDN \cite{sunehag2018value}}                                                           & \multicolumn{1}{c|}{15.56$\pm$19.02}         & \multicolumn{1}{c|}{84.10$\pm$8.42}   & \multicolumn{1}{c|}{65.05$\pm$2.15}           & \multicolumn{1}{c|}{99.38$\pm$1.25}     & \multicolumn{1}{c|}{85.88$\pm$6.02}             &  \multicolumn{1}{c|}{96.67$\pm$3.57}      \\ \hline
        \multicolumn{1}{|c|}{QMIX \cite{rashid2018qmix}}                                                           & \multicolumn{1}{c|}{0.32$\pm$0.64}         & \multicolumn{1}{c|}{84.28$\pm$1.87}   & \multicolumn{1}{c|}{72.33$\pm$10.03}           & \multicolumn{1}{c|}{99.07$\pm$1.25}     & \multicolumn{1}{c|}{97.92$\pm$1.77}             &  \multicolumn{1}{c|}{98.09$\pm$1.34}      \\ \hline
        \multicolumn{1}{|c|}{QTRAN \cite{son2019qtran}}                                                        & \multicolumn{1}{c|}{7.46$\pm$14.65}         & \multicolumn{1}{c|}{83.67$\pm$6.10}   & \multicolumn{1}{c|}{29.96$\pm$7.73}           & \multicolumn{1}{c|}{98.08$\pm$2.57}     & \multicolumn{1}{c|}{90.08$\pm$4.54}             &   \multicolumn{1}{c|}{97.53$\pm$1.51}     \\ \hline
        \multicolumn{1}{|c|}{QPLEX \cite{wang2021qplex}}                                                           & \multicolumn{1}{c|}{\textbf{86.25$\pm$2.25}}         & \multicolumn{1}{c|}{89.22$\pm$1.79}   & \multicolumn{1}{c|}{74.37$\pm$4.88}           & \multicolumn{1}{c|}{99.53$\pm$0.32}     & \multicolumn{1}{c|}{98.28$\pm$1.47}             &  \multicolumn{1}{c|}{97.97$\pm$0.60}      \\ \hline
        \multicolumn{1}{|c|}{DMIX \cite{sun2021dfac}}   &
        \multicolumn{1}{c|}{35.49$\pm$29.38}         & \multicolumn{1}{c|}{93.14$\pm$7.69}   & \multicolumn{1}{c|}{68.48$\pm$10.25}           & \multicolumn{1}{c|}{98.39$\pm$1.61}     & \multicolumn{1}{c|}{96.37$\pm$3.91}             &  \multicolumn{1}{c|}{96.71$\pm$3.13}      \\ \hline
        \multicolumn{1}{|c|}{DQMIX}                                         & \multicolumn{1}{c|}{64.95$\pm$8.35}         & \multicolumn{1}{c|}{\textbf{98.35$\pm$2.13}}   & \multicolumn{1}{c|}{\textbf{75.59$\pm$3.45}}           & \multicolumn{1}{c|}{\textbf{99.88$\pm$0.23}}     & \multicolumn{1}{c|}{\textbf{98.67$\pm$1.22}}            &   \multicolumn{1}{c|}{\textbf{99.07$\pm$1.20}}     \\ \hline
        \end{tabular}
        }
        \caption{The average values and standard deviations of test win rate (\%) of 5 independent runs after 20 million training timesteps of value-based MARL algorithms and variants for QMIX under distributional frameworks.}
        \label{tab:smac}
\end{table*}


We further conduct experiments on the SMAC \cite{samvelyan19smac} benchmark to evaluate:
(1) the performance of MCMARL compared to DFAC, in terms of the distributional framework;
(2) the performance of DQMIX (MCMARL variant of a recent model QMIX) compared to MARL baselines.


Before the discussion of the results, we briefly introduce experimental settings.
The experimental environment is the SMAC~\cite{samvelyan19smac} benchmark, which is based on the popular real-time strategy game StarCraft II.
The common hyperparameters of all methods are set to be the same as that in the default implementation of PyMARL.
For our MCMARL framework, we set $M=51$ (refer to C51 \cite{bellemare2017a}) and choose $V_{min}=-10$, $V_{max}=20$ from preliminary experiments on SMAC. 
To speed up the data collection, we use parallel runners to generate a total of 20 million timesteps data for each scenario and train the network with a batch of 32 episodes after collecting every 8 episodes. 
Performance is evaluated every 10000 timesteps with 32 test episodes. 

To compare the performance of MCMARL and DFAC, we test the variants of VDN and QMIX under the two distributional frameworks, as shown in Table~\ref{tab:smac_dis}.
DDN and DMIX are the DFAC variants of VDN and QMIX, respectively, while DVDN and DQMIX are our MCMARL variants of VDN and QMIX.
It can be observed that MCMARL variants consistently achieve better performance than DFAC variants, which demonstrates the superiority of our MCMARL framework in distributional MARL.
Besides, the learning curves of these methods in Figure~\ref{fig:smac} show that DQMIX outperforms the baselines with faster convergence.
\begin{figure}[t]
	\centering
	\includegraphics[width=0.45\textwidth]{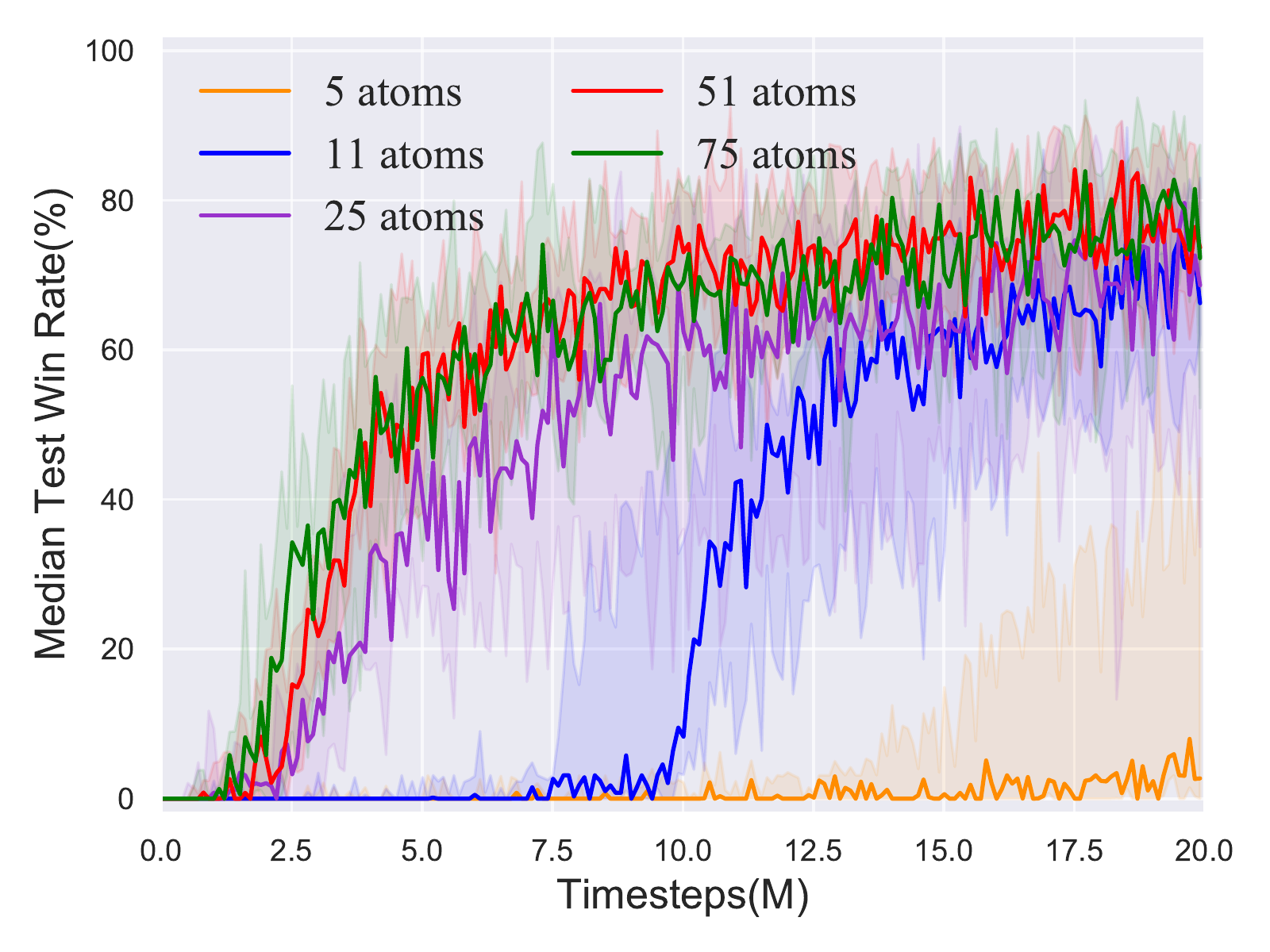}
	\caption{The test win rate curves of DQMIX on 5m\_vs\_6m with different number of atoms.}
	\label{fig:ablation}
\end{figure}

To illustrate the efficacy of DQMIX, we implement representative value-based MARL baselines.
The final performance of these algorithms is presented in Table~\ref{tab:smac}. 
We can see that DQMIX achieves the highest final average test win rate across almost all scenarios compared with the baseline algorithms, which indicates the advantage of approximating value functions via distribution. 
The outperformance of QPLEX over DQMIX on the corridor scenario might contribute to the attention mechanism.
In the future, we will design the multiplication operator on distributions to support the MCMARL variant of such attention-based models, which would be a fairer evaluation.

\subsection{Impact of Atom Number}
Furthermore, we conduct an ablation experiment to study MCMARL’s performance in relation to the number of atoms of categorical distribution, which is the core hyperparameter in our framework.
Figure~\ref{fig:ablation} reports the test win rate curves of DQMIX on 5m\_vs\_6m by varying the number of atoms with the value of $\{5, 11, 25, 51, 75\}$. 
It can be observed that the performance is extremely poor when the number of atoms is 5 and the test win rate reaches an acceptable performance when the number is greater than 11.
The results indicate that more atoms contribute to better performance and the performance approaches saturation as the number of atoms increases.
This is consistent with the fact that, given the fixed value range, support with more atoms has better expressive power and the learned distribution is more likely to be close to the true one. 
Considering that the performance is approaching saturation when the support size is greater than 51, we set the number of atoms to be $51$ throughout this work, to balance the effectiveness and efficiency.

\section{Conclusion}

In this paper, we propose MCMARL, a novel distributional value-based MARL framework, which explicitly models the stochastic in long-term returns by categorical distribution and enables to extend the existing value-based factorization MARL models to their distributional variants. 
To integrate the individual Q-value distributions into the global one, we design five basic distributional operations and theoretically prove that they satisfy the DIGM principle.
In this way, the MCMARL variants composed of these five basic operations meet the DIGM principle, which ensures the feasibility of decentralized execution.
Beyond that, empirical experiments on the stochastic matrix game and SMAC benchmark demonstrate the efficacy of MCMARL.
The limitation of this work is that we do not support the multiplication of the distributions, thereby our MCMARL framework cannot be applied to the attention-based MARL models.
In the future, we will further investigate other basic operations on the distribution, such as multiplication.

\bibliographystyle{unsrt}  
\bibliography{ref}  





\newpage
\appendix

\section{Appendix}

\subsection{DIGM Proof}
\label{appendix1}
To ensure the consistency between joint and individual greedy action selections, the distribution mixing network must satisfy the DIGM~\cite{sun2021dfac} condition, which is formulated as follows:
\begin{equation}
\begin{aligned}
    \arg\max_{\boldsymbol{a}} \mathbb{E} \left[ Z_{tot}(\boldsymbol{\tau}, \boldsymbol{a}) \right] = \left(
    \begin{matrix}
    \arg\max_{a_1} \mathbb{E} \left[ Z_1(\tau_1, a_1) \right] \\
    \vdots \\
    \arg\max_{a_N} \mathbb{E} \left[ Z_N(\tau_N, a_N) \right]
    \end{matrix}
    \right).
    \label{digm}
\end{aligned}
\end{equation}

One sufficient condition for the distribution mixing network that meets the DIGM condition is that all operations meet the DIGM condition.
The following propositions demonstrate that, under certain conditions, the five basic distribution operations, \emph{i.e.}, weighting, bias, convolution, projection and active function, satisfy the DIGM condition. We provide detailed proof for these propositions in this section.

\noindent\textbf{Proposition 1.} If $w \ge 0$, then
\begin{equation}
\arg\max_{a} \mathbb{E}[X(a))]= \arg\max_{a} \mathbb{E}[w \cdot X(a)].
\nonumber
\end{equation}
\begin{proof}[Proof]
\begin{align*}
  \arg\max_{a} \mathbb{E}[w \cdot X(a)] 
= \arg\max_{a} \sum_j p_j(w x_j)
\overset{(*)}{=} \arg\max_{a} \sum_j p_j x_j
= \arg\max_{a} \mathbb{E}[X(a))],
\end{align*}
where (*) is satisfied because of the monotonicity of function $f$.
\end{proof}

\noindent\textbf{Proposition 2.}
\begin{equation}
\arg\max_{a} \mathbb{E}[X(a)]= \arg\max_{a} \mathbb{E}[X(a) + b].
\nonumber
\end{equation}
\begin{proof}[Proof]
\begin{align*}
\arg\max_{a} \mathbb{E}[X(a) + b] 
= \arg\max_{a} \sum_j p_j(x_j+b)
= \arg\max_{a} \sum_j p_j x_j
= \arg\max_{a} \mathbb{E}[X(a))].
\end{align*}
\end{proof}

\noindent\textbf{Proposition 3.} 
\begin{equation}
\arg\max_{a_1,a_2} \mathbb{E}[X_1(a_1)*X_2(a_2)] =  
\left(
\begin{matrix}
\arg\max \limits_{a_1} \mathbb{E}[X_1(a_1)]\\
\arg\max \limits_{a_2} \mathbb{E}[X_2(a_2)]\\
\end{matrix}
\right).
\nonumber
\end{equation}

\begin{proof}[Proof]
\begin{equation}
\begin{aligned}
  &\mathbb{E}[X_1(a_1)*X_2(a_2)] \\\nonumber
  = & \sum_{i=0}^{M_1-1} \sum_{j=0}^{M_2-1} p_{1,i} p_{2,j} (x_{1,i} + x_{2,j}) \\\nonumber
  = & \sum_{i=0}^{M_1-1} p_{1,i} \left(\sum_{j=0}^{M_2-1} p_{2,j} x_{1,i} + \sum_{j=0}^{M_2-1} p_{2,j} x_{2,j}\right) \\\nonumber
  = & \sum_{i=0}^{M_1-1} p_{1,i} \Big(x_{1,i} + \mathbb{E}[X_2(a_2)] \Big) \\\nonumber
  = & \sum_{i=0}^{M_1-1} p_{1,i} x_{1,i} + \mathbb{E}[X_2(a_2)] \\\nonumber
  = & \mathbb{E}[X_1(a_1)]+\mathbb{E}[X_2(a_2)].
\end{aligned}
\end{equation}
Given the above equation,  
\begin{align*}
 \arg\max_{a_1,a_2} \mathbb{E}[X_1(a_1)*X_2(a_2)] 
=  \arg\max_{a_1,a_2} (\mathbb{E}[X_1(a_1)]+\mathbb{E}[X_2(a_2)])
= 
\left(
\begin{matrix}
\arg\max \limits_{a_1} \mathbb{E}[X_1(a_1)]\\
\arg\max \limits_{a_2} \mathbb{E}[X_2(a_2)]\\
\end{matrix}    
\right).
\end{align*}
\end{proof}

\noindent\textbf{Proposition 4.} For any atoms $[x]$,
\begin{equation}
\arg\max_{a} \mathbb{E}[X(a)]= \arg\max_{a} \mathbb{E}[\Phi_{[x]}(X(a))].
\nonumber
\end{equation}
\begin{proof}[Proof]
Considering projecting random variable distribution of $[x_j]$ to atoms $[\hat{x}_k]$, let's assume that the atom range after projection $[\hat{V}_{min},\hat{V}_{max}]$ is enough to cover all atoms  before projection, \emph{i.e.,} $\forall\  0\le j<M$, $x_j \in [\hat{V}_{min},\hat{V}_{max}]$.

$\forall\  x_j$, $\exists \  1 \leq k_j < K $, $s.t.$ $\hat{x}_{k_j-1} \le x_j < \hat{x}_{k_j}$, \emph{i.e.,} the immediate neighbours of $x_j$ is $\hat{x}_{k_j-1} $ and $ \hat{x}_{k_j}$. 
According to the definition of projection operation, the  probability  $p_j$ in $x_j$ is disassembled as  $\frac{\hat{x}_{k_j}-x_j}{\triangle \hat{x}} p_j$ in $\hat{x}_{k_j-1}$ and $\frac{x_j-\hat{x}_{k_j-1}}{\triangle \hat{x}} p_j$ in $\hat{x}_{k_j}$. 

\begin{equation}
\begin{aligned}
& \mathbb{E}[\Phi_{[x]}(X(a))]\\
=& \sum_k (\sum_j\left[ 1- \frac{|[x_j]^{\hat{V}_{max}}_{\hat{V}_{min}}-\hat{x}_k|}{\triangle \hat{x}} \right]^{1}_{0} p_j)\hat{x}_{k}\\
=&\sum_j(\sum_k \left[ 1- \frac{|[x_j]^{\hat{V}_{max}}_{\hat{V}_{min}}-\hat{x}_k|}{\triangle \hat{x}} \right]^{1}_{0} p_j\hat{x}_{k})\\
=& \sum_j(\frac{\hat{x}_{k_j}-x_j}{\triangle \hat{x}} p_j \cdot\hat{x}_{k_j-1} + \frac{x_j-\hat{x}_{k_j-1}}{\triangle \hat{x}} p_j \cdot \hat{x}_{k_j})\\
=& \sum_j\Big(\frac{p_j}{\triangle \hat{x}}[(\hat{x}_{k_j}-x_j)\cdot(\hat{x}_{k_j}-\triangle \hat{x}) 
\\& + (x_j-\hat{x}_{k_j}+\triangle \hat{x})\cdot\hat{x}_{k_j}]\Big)\\
=& \sum_j p_j x_j \\
= & \mathbb{E}[X(a)]
\end{aligned}
\end{equation}
Therefore,
\begin{equation}
\begin{aligned}
\arg\max_{a} \mathbb{E}[\Phi_{[x]}(X(a))]
= \arg\max_{a} \mathbb{E}[X(a)]\nonumber.
\end{aligned}
\end{equation}
\end{proof}

\noindent\textbf{Proposition 5.} For any monotone increasing function  $f:\mathbb{R} \rightarrow \mathbb{R}$, 
\begin{equation}
\arg\max_{a} \mathbb{E}[X(a))]= \arg\max_{a} \mathbb{E}[f (X(a))].
\nonumber
\end{equation}
\begin{proof}[Proof]
\begin{align*}
\arg\max_{a} \mathbb{E}[f (X(a))] 
= \arg\max_{a} \sum_j p_j(f( x_j))
\overset{(*)}{=} \arg\max_{a} \sum_j p_j x_j
= \arg\max_{a} \mathbb{E}[X(a)],
\end{align*}
where (*) is satisfied because of the monotonicity of function $f$.
\end{proof}

\subsection{Why DVDN performs unsatisfactorily}
\label{appendix2}
According to the network structure of DVDN, the effectiveness of DVDN relies on a strong assumption that the randomness of individual Q-value distributions can be directly summed up, which is inconsistent with the most cases in reality.
Take a simple game as an example to illustrate the limitation of DVDN.
Suppose that there exists a system with two agents, $A$ and $B$, the reward they get from the environment is stochastic and relative. For example, there is a 50\% possibility that $A$ is rewarded with 1 while $B$ is rewarded with -1 and another 50\% chance that $A$ gains -1 while $B$ gains 1. In this case, after convolution of DVDN, the model will think that the system has a 25\% chance of getting a reward of 2, 25\% chance of getting a reward of -2 and the rest to gain 0. But in fact, we can know that the expectation reward of the system is 0 for 100\%. This example reveals that stochasticity of the environment can not be added simply so that direct summation after convolution will result in distortion during learning. In order to precisely approximate the return of the environment, it's necessary to make use of global state to introduce nonlinearity to the training process. To this end, DQMIX under our MCMARL framework manages to achieve a satisfactory performance in the experiments while DVDN performs poorly. To be mentioned, as the distributional variant of VDN in DFAC, \emph{i.e.} DDN, also makes approximation by direct summation, it performs even worse than our DVDN, proving the effectiveness and robustness of our framework.


\end{document}